\documentclass[preprint,12pt]{elsarticle}

\usepackage{amssymb}

\usepackage{times}
\usepackage{epsfig}
\usepackage{graphicx}
\usepackage{subcaption}
\usepackage{amsmath}
\usepackage{amsthm}
\usepackage{amssymb}
\usepackage{amsfonts}
\usepackage[breaklinks=true,bookmarks=false]{hyperref}
\usepackage{booktabs}
\usepackage{tabu}
\usepackage{color,soul}
\sethlcolor{cyan}
\usepackage{xcolor,colortbl}
\definecolor{Gray}{gray}{0.85}
\newcolumntype{g}{>{\columncolor{Gray}}l}

\usepackage{tabularx}
\usepackage{multirow, makecell}
\newcolumntype{C}{>{\centering\arraybackslash}X}
\newcolumntype{L}{>{\raggedright\arraybackslash}X}
\newcolumntype{R}{>{\raggedleft\arraybackslash}X}
\journal{Pattern Recognition}

\begin{document}

\begin{frontmatter}



\title{Object Tracking and Geo-localization from Street Images}


\author[inst1]{Daniel Wilson \thanks{These authors contributed equally.}}
\author[inst1]{Thayer Alshaabi \footnotemark[1]}
\author[inst1]{Colin Van Oort \footnotemark[1]}
\author[inst1]{Xiaohan Zhang}
\author[inst2]{Jonathan Nelson}
\author[inst1]{Safwan Wshah}

\affiliation[inst1]{organization={Vermont Complex Systems Center},
            addressline={82 University Pl}, 
            city={Burlington},
            postcode={05405}, 
            state={Vermont},
            country={United States}}

\affiliation[inst2]{organization={Vermont Agency of Transportation},
            addressline={133 State St}, 
            city={Montpelier},
            postcode={05633}, 
            state={Vermont},
            country={United States}}

\begin{abstract}
Geo-localizing static objects from street images is challenging but also very important for road asset mapping and autonomous driving. In
this paper we present a two-stage framework that detects and geolocalizes traffic signs from low frame rate street videos.
Our proposed system uses a modified version of RetinaNet (GPS-RetinaNet), which predicts a positional offset for each sign relative to the camera, in addition to performing the standard classification and bounding box regression.
Candidate sign detections from GPS-RetinaNet are condensed into geolocalized signs by our custom tracker, which consists of a learned metric network and a variant of the Hungarian Algorithm. 
Our metric network estimates the similarity between pairs of detections, then the Hungarian Algorithm matches detections across images using the similarity scores provided by the metric network.
Our models were trained using an updated version of the ARTS dataset, which contains 25,544 images and 47.589 sign annotations ~\cite{arts}.
The proposed dataset covers a diverse set of environments gathered from a broad selection of roads.
Each annotaiton contains a sign class label, its geospatial location, an assembly label, a side of road indicator, and unique identifiers that aid in the evaluation. 
This dataset will support future progress in the field, and the proposed system demonstrates how to take advantage of some of the unique characteristics of a realistic geolocalization dataset.
\end{abstract}


\begin{highlights}
\item A large and realistic dataset  to support research in the field of object geolocalization
\item An object detector designed to predict GPS locations using a local offset and coordinate transform
\item A tracker which condenses object detections into geolocalized predictions
\end{highlights}


\end{frontmatter}


\section{Introduction} 
\label{Introduction}

Recent years have been marked by improved performance in deep learning and its application to an increasingly broad set of domains.
Object geolocalization is particularly crucial for creating and maintaining large-scale geospatial maps of road assets and autonomous driving systems.
Few pipelines applied to geolocalization datasets contain subcomponents which perform object detection and tracking~\cite{uber}, before producing geolocalized objects in the final stage.
Object detection algorithms can identify and classify an unknown number of objects present in an image, assigning a label to each detected object and drawing a bounding box around it.
Object tracking can then be used to pair detected objects across sequential frames, such that detections for the same object are joined together across all frames in which they appear.
Object geolocalization is the process of taking objects identified in one or more images and determining their geospatial location represented as GPS coordinates.

To train our system and support future research we constructed the ARTS v2 dataset, which builds on the ARTS dataset~\cite{arts} to become the largest dataset of geolocalized US traffic signs.
ARTS v2 contains US traffic signs with class labels consistent with the Vermont VCI sign catalog~\cite{vci_signs} and contains 25,544 images, 47,589 annotations, and 199 sign classes.
ARTS v2 extends the annotation information present in ARTS to include a sign assembly indicator, a side of road indicator, and a unique sign identifier that aids in the evaluation of object tracking algorithms.

Our proposed geolocalization pipeline is a two-stage system.
The first stage of our system is our object detector called GPS-Retinanet.
We have developed GPS-Retinanet by augmenting the popular object detector Retinanet~\cite{lin2018retinanet}.
Our detector contains multiple innovations, including a new loss function modified from Focal Loss to better cope with class imbalance present in the ARTS dataset Figure~\ref{fig:loss}, and a GPS-Subnet along with a coordinate transform which enables it to predict GPS coordinates using the architecture displayed in Figure ~\ref{fig:gps-retinanet}.
The second stage of our system is a learned heuristic that computes what can be intuitively interpreted as a similarity metric between two detections.
We accomplish this by training a siamese neural network which receives features from a pair of detections as input.
The network learns to predict whether the input detections refer to the same or a different physical sign, which serves as a measure of similarity between the two detections.
Then, our system employs the Hungarian algorithm~\cite{hungarian} to pair detections computed to have greater similarity, while splitting detections of lesser similarity that appear to belong to different signs.
After applying this algorithm while sequentially stepping through images, it produces lists of detections estimated to belong to the same physical sign known as tracklets.
Each tracklet is condensed into a single geolocalized sign prediction using a weighted average of the tracklet components.

In this paper, our goal is to geolocalize road assets, starting with traffic signs.
In pursuit of this, we develop three research contributions.
First, we provide an enhanced version of the ARTS~\cite{arts} dataset, ARTS v2, to serve as a benchmark for object geolocalization. 
Second, we provide an object detection system capable of predicting geolocations using only images and camera pose information.
Third, we employ a tracker to collapse a set of detections in a noisy, low-frame rate environment into final geolocalized asset predictions.

\section{Related Work} 
\label{Related-Work}
Object Geolocalization from street images has been the focus of important recent research. 
Before deep learning, a common approach for object geolocalization was to use an epipolar constraints~\cite{szeliski2010computer} to reconstruct 3D points from corresponding image locations. 
\cite{fairfield2011traffic} used this method to predict traffic light locations. 
\cite{SOHEILIAN20131} used it to triangulate and estimate the locations of traffic signs that were detected from their silhouette. 
\cite{hebbalaguppe2017telecom} proposed a pipeline for telecom inventory management by triangulating telecom assets which are detected using a histogram of oriented gradients (HOG) feature descriptor with a linear SVM~\cite{dalal2005histograms} from Google Street View (GSV) images. 
While the methods discussed above may be suitable for some applications, 
their accuracy and robustness are bounded by the specific technical limitations of the techniques used for feature detection and point matching~\cite{fairfield2011traffic,hebbalaguppe2017telecom}, 
and often require prior domain knowledge of the target objects~\cite{SOHEILIAN20131}.  

Deep neural networks (DNNs) are increasingly becoming the state-of-the art technique in geolocalization. 
\cite{Krylov1} proposed a framework that integrates Markov Random Field (MRF) and monocular depth estimation using a convolutional neural network (CNN) for object geolocalization.
They later expanded their method, incorporating point cloud data captured from drones to further improve geolocalization accuracy \cite{Krylov2}. 
Each object is assumed to have different geolocation, which fails on objects such as assembled signs. 
\cite{LOB} proposed a line-of-bearing based geolocalization framework to estimate the location of electric poles in GSV images which is detected by a CNN object detector. 
\cite{nassar1} combined object detection and object re-identification in a jointly learning task by applying a soft geometric constraint on detected objects from GSV images. 
Recently, \cite{nassar2} proposed GeoGraph, a Graph Neural Network (GNN) based method for geolocalization. 
By leveraging the geometric cues, the GNN learns to match features of the same objects across frames and predict positions.

Similar to our approach, \cite{chaabane2021end} proposed an end-to-end trainable object geolocalizaton and tracking model composed of an object pose regression network and object matching network. 
Their method, however, requires an accurate camera intrinsic matrix as input and their system needs the objects' 3D information for training which is annotated from Li-DAR data by human experts. 
Finally, as mentioned in \cite{chaabane2021end}, only scenes in which a traffic light appears in at least 5 unique key frames are be selected to in the dataset. 
However, this requirement is not representative of practical street environments. 

By contrast, our proposed algorithms can handle a large quantity of objects that may appear only once in a sequence. 
Our approach works for most street image datasets in which sequences of frontal images are available without camera calibration or object rotation information, which are rarely available.

Many tracking-by-detection frameworks have been developed over the past decade for a wide range of applications~\cite{lin2017fpn,girshick2014rcnn,girshick2015fastrcnn,ren2015fasterrcnn,lin2018retinanet,he2017maskrcnn,lin2017fpn}.
Unlike some existing algorithms that use visual cues and motion tracking to trace objects in a sequence of images~\cite{zhu2018online,voigtlaender2019mots,son2017multi,xu2019spatial}, we train a network to estimate a similarity score for a given pair of objects, which we can use to map objects across consecutive frames.  
We directly input the predicted object location, camera location, camera heading, and predicted bounding box information into the similarity network.

Other research has developed object trackers using a similar premise. 
\cite{similarity_tracker} use a deep siamese convolutional network to learn a similarity function trained during an offline learning phase, which is then evaluated during tracking. 
\cite{MOT_MDP} show that multiple object tracking can be modeled using a Markov decision process (MDP). 
Researchers show that it is possible to use dual matching attention networks, incorporating both spatial and temporal information to generate attention maps on input images to perform tracking~\cite{dual_attention}. 
Most approaches are designed to only use visual features as the vast majority of publicly available datasets do not provide any further metadata.

\section{Datasets} 
\label{arts}

\begin{table*}[!tp] 
    \tiny
    \centering 
    \begin{tabu} to \linewidth {l C C cc C} 
        \toprule 
        & \textbf{Pasadena Multi-View ReID}~\cite{nassar1} 
        & \textbf{Traffic Light Geolocalization (TLG)}~\cite{uber}
        & \multicolumn{2}{c}{\textbf{ARTS v1.0}~\cite{arts}} 
        & \textbf{ARTS v2.0} \\ 
        & 
        &
        & \textsc{Easy} & \textsc{Challenging} 
        &  \\ \midrule \midrule 
        Number of classes       & 1 & 1 & 62 & 175 & 199 \\ \midrule
        Number of images        & 6141 & 96960 & 6807 & 16023 &  25544 \\ \midrule
        Number of annotations   & 25061 & Unknown & 9006 & 27181  & 47589 \\ \midrule
        Side of the road        & & &  &  &  \checkmark \\ \midrule
        Assembly                & & &  &  &  \checkmark \\ \midrule
        Unique Object IDs       & \checkmark & \checkmark &  &  &  \checkmark \\ \midrule
        5D Poses                & & \checkmark & & & \\ \midrule
        GPS                     & \checkmark & \checkmark & \checkmark  & \checkmark &  \checkmark \\ \midrule
        Color Channels          & RGB & RGB & RGB & RGB  & RGB \\ \midrule
        Image Resolution        & $2048\times1024$ & $1600\times1900$ & $1920\times1080$ & $1920\times1080$ & $1920\times1080$ \\ \midrule
        Publicly Available      & & \checkmark & \checkmark & \checkmark & \checkmark \\ \bottomrule
    \end{tabu}
    \caption{A comparison between the Pasasena multi-view object reidentification~\cite{nassar1}, the traffic light geolocalization (TLG)~\cite{uber}, ARTS v1.0 easy and challenging, and ARTS v2.0 datasets.
    } 
    \label{tab:dataset-stats} 
\end{table*}

Despite recent interest, there are only a couple of datasets proposed to support reasearch in object geolocalization.
\cite{nassar1} proposed a multi-view dataset in which the goal is to re-identify streetside trees from different views.
It includes 6020 individual trees, 6141 GSV image formatted as panoramas, and 25061 bounding boxes.
Each tree is annotated in its four closest panoramas, and is labeled with a unique ID so re-identification can be performed.
However, their dataset is not publicly available.
Researchers from Uber~\cite{uber} compiled another dataset for traffic light detection derived from nuScenes---a popular open-source dataset for autonomous driving~\cite{caesar2020nuscenes}.
The dataset has 400 scenes, each lasting 20 seconds with 12 frames per second.
All images have metadata indicating the 5D pose of the camera and each annotated traffic light.
Each traffic light can be distinguished with a uniquely assigned ID.

Here, we present an expansion to the ARTS dataset~\cite{arts}, providing new metadata to enable future research in this domain. 
The original ARTS dataset is composed of over 16,000 images spanning 175 different classes.
The dataset is structured into sequences of images referred to as road segments.
Each segment contains a sequence of images taken from a camera mounted to the top of a car driving down a road, with roughly one second intervals between each image due to the need to store the dataset in a reasonable amount of space.
Each image contains annotations for each readable sign, and each annotation specifies a bounding box around the sign and its class, along with the GPS coordinates of that sign.
The camera's coordinates and heading are also available for each image.
The ARTS dataset is presented in three different configurations: easy, challenging, and video-logs.
All three configurations of the dataset provide manually-labeled annotations that are similar to the PASCAL VOC format~\cite{pascal}. 
The easy version of the dataset contains a total of $\sim$10K images and $\sim$17K annotations, covering 78 different sign classes.
All annotated signs in the easy version were captured within a 30 meter radius of the camera with a minimum of 50 samples per class.
The challenging version of the dataset contains a total of $\sim$35K annotations scattered in $\sim$20K images, covering 171 sign classes, with a minimum of 20 samples per class captured from a distance up to 100 meters.All of our experiments in this paper were implemented and evaluated on the challenging version. 

Substantial enhancements have been made to the second iteration of the ARTS dataset, increasing the number of images to 25544, the number of unique sign classes to 199, and the number of annotations to 47589.
Moreover, each sign annotation has been updated with additional attributes.
First, each annotation specifies the 'sign side', which indicates the side of the road the sign is on, labeled as either left or right.
We note that the sign side label indicates the positional location of the sign relative to the road not to the camera.
Second, each annotation has a binary attribute marking whether the sign is a part of an assembly---a group of signs supported by the same post.
All the signs in an assembly will have this boolean attribute set to True, whereas individual signs are set to false. Even though the assembly identifier is useful, we didn't use it for our proposed research, instead leaving it for future research.

Finally, each physical sign in a road segment has been given unique integer identifier.
Since most signs appear in multiple images, a sign annotation will have the same ID each time the same physical signs appear.

Compared to other traffic recognition and and geolocalization datasets, ARTS is the largest in terms of both number of images, classes, and annotations.
The dataset contains high quality 1920 x 1080 resolution images, available in multiple formats including video logs and individual annotations in a format similar to the PASCAL VOC format.
ARTS v2 is also the only dataset containing labels specifying the side of road and assembly attributes.
Table~\ref{tab:dataset-stats} shows a full comparison between ARTS and similar geolocalization datasets in terms of number of classes, number of images, and number of annotations for each dataset.

Current datasets for object geolocalization algorithms are simple and constructed under ideal circumstances~\cite{nassar1}~\cite{uber}.
The ARTS v2 dataset contains multiple unique challenges which makes it more representative of the task encountered in the real world.
First, ARTS v2 features 199 different sign classes appearing with a highly imbalanced distribution, thereby classes such as stop signs appear far more frequently than more obscure classes of signs.
The heavy-tailed distribution increases the difficulty of training models to predict sign class since less frequent classes have few training samples, as is representative of what we expect to see in the real world.

US traffic sign classification is particularly challenging due to the inconsistency between states.
While the US Department of Transportation standards are followed to varying degrees, there are a wide variety of specific traffic sign configurations across states’ road networks.
Roads contain many signs that do not conform to known standards.

There are also many objects with similar appearance to road signs, which tends to lead to false positives from object detectors.
Business signs and billboards, hand-made signs placed for events like yard sales, and car license plates all have a tendency to fool detection models.

Due to data storage constraints, Department of Transportation collected street image data and stored it at a low frame rate (~ 1 frame per sec) resulting in ~200-400 images per mile.

Another challenge is that, as with any dataset, there may exist objects with missing annotations.
Other similar scenarios are signs that are rotated at an angle relative to the camera, or signs that are partially obscured by other objects such as trees.
Annotators were asked to use their best judgement for annotated signs a detection model should be expected to detect.

Furthermore, our dataset resembles a wide variety of driving environments.
There are road segments corresponding to highways, small rural roads, complex intersections, and busy city roads.
The car travels at a variety of speeds and takes many turns, making the challenge of associating detections between frames far more difficult.
Finally, sign assemblies are particularly challenging for several reasons. 
Not only do assemblies contain many signs which need to be individually detected and geolocalized on a single post, but signs of similar appearance are particularly likely to be grouped on the same assembly.
For example, assemblies containing multiple signs each of which indicates a highway are common.
All of these signs on such an assembly would have the same GPS coordinates, and many would have the same class and often extremely similar appearance.
Distinguishing between these signs when collapsing the set of detections into separate geolocalized sign predictions is challenging because of how many characteristics are shared amongst physically separate signs.

We have made the ARTS v2 dataset publicly available to support research and development in both traffic sign recognition and object geolocalization.
ARTS v2 can be accessed at the following \href{https://drive.google.com/drive/u/1/folders/1u_nx38M0_owB0cR-qA6IOWgZhGpb9sWU}{link}.

\section{Methods}
\label{Methods}

\begin{figure*}[tp!]
    \centering
    \includegraphics[width=\linewidth,keepaspectratio]{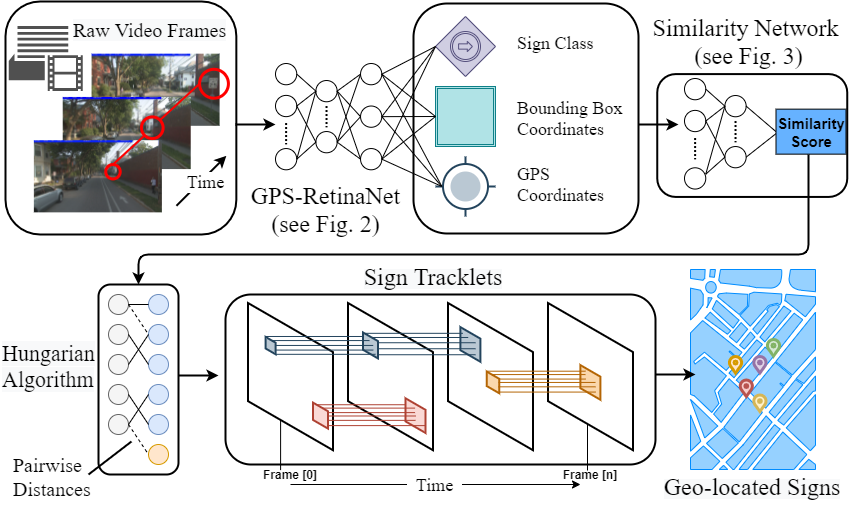}
    \caption{An overview of the Sign Hunter pipeline.}
    \label{fig:pipeline}
\end{figure*}

\subsection{GPS RetinaNet} 
\begin{figure*}[tp!]
    \includegraphics[width=\linewidth,keepaspectratio]{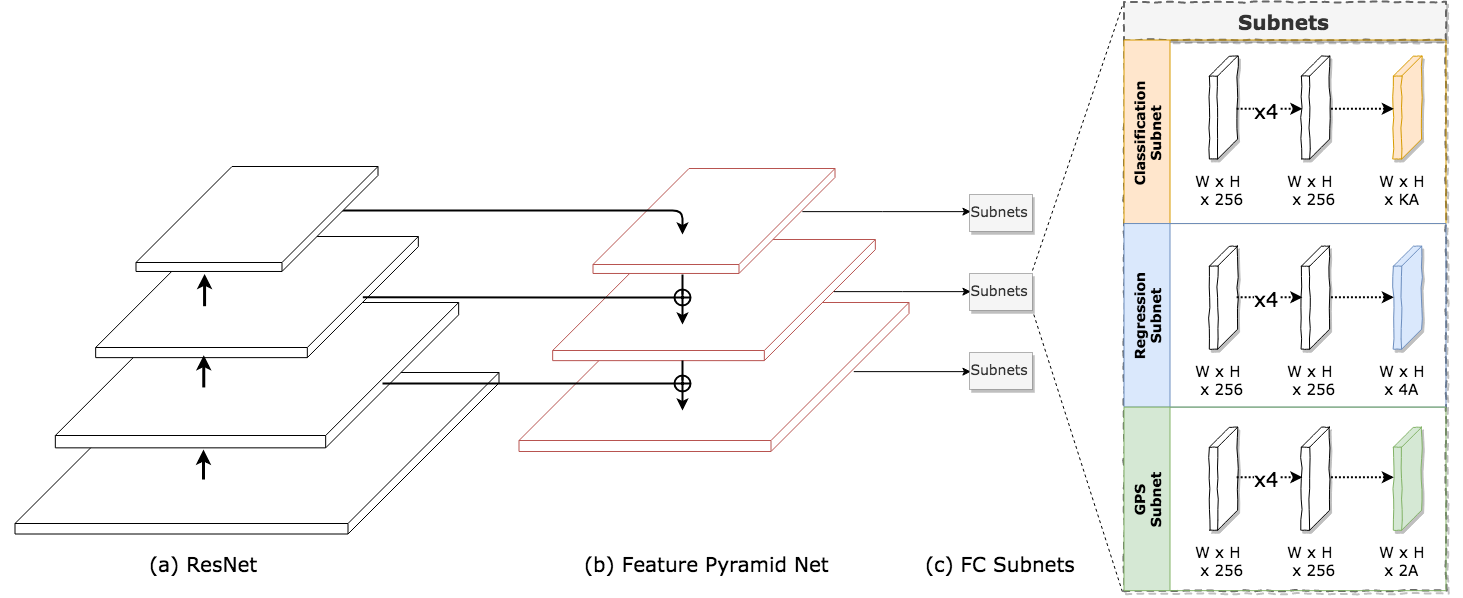}
    \caption{
        \textbf{GPS-RetinaNet.}
        Similar to RetinaNet~\cite{lin2018retinanet}, this architecture uses a FPN~\cite{lin2017fpn} backbone on top of a ResNet~\cite{he2016resnet} model \textbf{(a)} to create a convolutional feature pyramid \textbf{(b)}.
        Then, we attach 3 sub-networks \textbf{(c)}; one for classification, one for box regression, and one for GPS/depth regression.
    } 
    \label{fig:gps-retinanet} 
\end{figure*}

Estimating absolute distance of relative objects using vision-based systems is very challenging, especially with the hardware limitations imposed by single-vision cameras.
It becomes even more difficult when objects of similar appearance have different sizes as is the case in ARTS v2.
Most existing models rely on sensors like RADAR or LIDAR to estimate distances to surrounding objects.
Our model relies solely on a single-vision camera to predict the GPS coordinates of an object on the road.
We add a fully connected network which extracts features from RetinaNet's feature pyramid~\cite{lin2018retinanet} which learns to regress a detected object's offset in a coordinate system local to the image.
We refer to this addition as the GPS-subnet, which expands RetinaNet's base architecture as displayed in Figure~\ref{fig:gps-retinanet}.
The architecture of all three sub-networks: classification, box-regression, and GPS-regression, is identical except for the last layer in each sub-network.
The classification sub-network terminates with a ($K \times A$) linear outputs, where $A$ is the number of different anchors used in the network and $K$ represents the number of classes.
The box-regression sub-network ends with a ($4 \times A$) linear outputs to determine the relative position of the object~\cite{lin2018retinanet}.
The GPS sub-network concludes with a ($2 \times A$) linear outputs for each spatial level in the network.

More specifically, the GPS-subnet learns to regress two local offset values, one vertical and the other horizontal, between the camera's position and the position of each detected object.
The local vertical and horizontal offset values are measured in meters.
These offsets are then fed into a coordinate transform to generate the object's predicted GPS as follows:
\begin{align}
    X_r &= X_o * \cos{\theta} + Y_o * \sin{\theta} \\
    Y_r &= X_o * \sin{\theta} - Y_o * \cos{\theta} \\
    O_{lat} &= X_r / 6378137 \\
    O_{lon} &= Y_r / (6378137 * \cos(\pi * C_{lat} / 180)) \\
    P_{lat} &= C_{lat} + O_{lat} * 180 / \pi \\
    P_{lon} &= C_{lon} + O{lon} * 180 / \pi
\end{align}
The variables $X_o$ and $Y_o$ represent the respective horizontal and vertical offsets predicted by the network from the perspective of the image in meters. 
We use $\theta$ to represent the camera's facing direction (measured with a compass), and $C_{lat}$ and $C_{lon}$ indicate the camera's latitude and longitude. 
Both $X_r$ and $Y_r$ are calculated as the meters offsets and rotated onto the latitudinal and longitudinal coordinate system. 
Hence, $O_{lat}$ and $O_{lon}$ are offsets converted from meters to latitude and longitude, 
and $P_{lat}$ and $P_{lon}$ provide the final latitude and longitude prediction of the detected object after adding the predicted offset to the camera coordinates.
Objects in the ARTS dataset are annotated according to their GPS coordinates, which means that we must convert these coordinates to local image offsets in meters to provide supervision for the GPS sub-network during training.

To handle the class imbalance present in ARTS v2, we propose a modification to Focal Loss that replaces $\gamma$ in the original definition by an adaptive modulator. 
We define the new focusing parameter as: 
\begin{align}
    \Gamma & = e^{(1 - p_t)}, \\
    \textnormal{FL}e(p_t) & = -(1 - p_t)^{\Gamma} \log(p_t).
\end{align} 

For convenience, we will refer to our new definition of Focal Loss as (FL$e$) throughout the paper.
FL$e$ introduces two new properties to the original definition.
It dynamically fine-tunes the exponent based on the given class performance to reduce the relative loss for well-classified classes maintaining the primary benefit of the original FL.
Figure~\ref{fig:loss} directly compares FL with FL$e$, highlighting that FL$e$ (shown in green) crosses over $FL_{\gamma=2}$ (shown in orange) around ($p_t = .3$).
As $p_t$ goes up from $.3 \to 1$, FL$e$ starts to shift up slowly ranging in between FL and Cross Entropy CE (shown in blue).
See Appendix~\ref{appendix} for more technical details. 

\subsection{Similarity Network} 
When GPS-RetinaNet is applied to an image, it produces detections for each sign specifying a bounding box, sign class, and (after a coordinate transform) GPS coordinates.
Images in the ARTS dataset have a spacing of around 8 meters, meaning the same sign is likely to appear and be detected in multiple frames.
Since our final goal is to produce a set of geolocalized sign predictions, we need to collapse the multiple detections produced for many signs into a single prediction for each distinct, physical sign.
Our proposed solution is a tracker that iteratively steps through the images in each road segment from the ARTS v2 video logs, and merges multiple detections from the same sign appearing in different frames and splitting detections belonging to different signs.
The tracker is composed of two core components, the distance metric network and the Hungarian algorithm.
The role of the first core component of our tracker is to compute a learned heuristic indicating how likely it is a pair of detections provided by GPS-RetinaNet refer to the same sign.
This task is handled by a neural network we refer to as the similarity network.

The similarity network receives as input all useful information produced by the object detector, so it can learn to distinguish whether two detections in different frames correspond to the same or different signs.
The network processes these inputs using two siamese sub-network and a third sub-network designed to handle the remaining inputs.
For each detection, the similarity network receives the car's GPS and heading, the sign class, GPS, and bounding box predicted by the object detector, and image pixels within the bounding box as inputs.
The pixel information for the two signs are resized from their natural resolution to $32 \times 32 \times 3$ using bi-linear interpolation.
The final inputs to the network are rank 3 tensors containing a ``snapshot'' of basic information from of all the detections produced by RetinaNet in the two frames the detections originate from.
Each detection in a frame is assigned to a square in a $10 \times 10$ grid corresponding to the location at which the object was detected in the image.
The depth axis of the tensor contains a vector of features from the detection, including its predicted class and GPS coordinates.
This forms a tensor containing information from all of the detections in the image, not just the two being compared, to allow the network to perceive the position of the detections of interest relative to other detected objects.

The network is trained to predict a value of either 0 to 1, as an estimate of the probability that a pair of detections belong to the same sign.
These values can be interpreted as a distance metric, where output values closer to 0 indicate the sign detections have greater similarity, and are thus less distant from one another.
Conversely, detections from different signs should have outputs closer to 1, indicating that the detections are more distant from one another.

\begin{figure*}[tp!]
    \centering
    \includegraphics[width=0.8\textwidth,keepaspectratio]{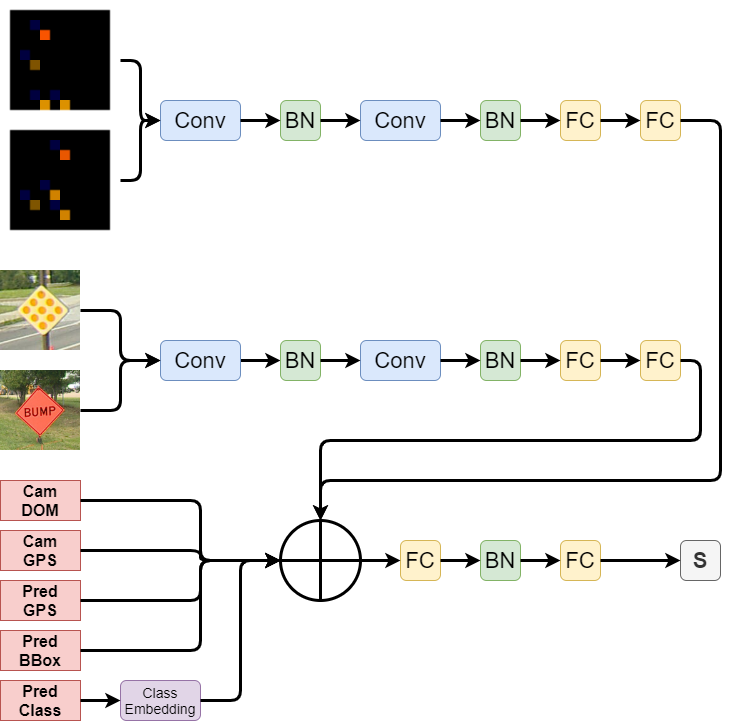}
    \caption{
        The architecture of the similarity network. The network uses features provided by the object detector as inputs to compute similarity scores between pairs of detections.
    }
    \label{fig:similarity_network}
\end{figure*}

The predicted sign class is converted to a vector of length 50 using an embedding layer.
Camera heading and GPS, and the predicted sign GPS and bounding box are all concatenated together along with the embedding layer.
Meanwhile, the $32 \times 32 \times 3$ scaled images containing the pixels from the detections are fed through a siamese sub-network consisting of multiple convolutional layers and then a few fully connected layers resulting in a vector of 32 features.
The 3D tensors containing a snapshot of all the detections from the two frames is processed by a second siamese sub-network consisting of multiple convolutional layers and fully connected layers resulting in a vector of 8 features.
All the aforementioned vectors are concatenated together and sent through multiple fully connected layers to generate the final prediction.
The exact architecture is shown in Figure~\ref{fig:similarity_network}.
We train this metric network for 20 epochs using the binary crossentropy loss function and the Adam optimizer with an initial learning rate of 0.01.

The final task in training the similarity network is to develop a satisfactory noise distribution when training.
The key challenge is that during training the similarity network needs binary labels indicating if each pair of two detections produced by RetinaNet are the same or not, but such labels only exist for annotated data.
We therefore train the distance network to estimate if two annotations (as opposed to detections) are the same, but with the caveat that extreme care should be taken to add noise to the annotations that accurately reproduces the behavior of the object detector, since we found the network's performance to be very sensitive to discrepancies between annotated and detected objects.
To accomplish this, we implemented a simple algorithm which tests if an annotation has an obvious match predicted by RetinaNet, by checking to see if the annotation's bounding box has an obvious match 
(indicated by an $\textnormal{IOU} > 0.9$) with one and only one detection.
If this is the case, we measure the latitude and longitude discrepancy between the annotation and detection, create a boolean variable indicating if the classes match, and measure the change in the bounding box.
By repeating this process for each annotation, we construct a noise distribution representing how often and by how much the detections differ from the annotations.
When then which we sample from this noise distribution when training our similarity network.

\subsection{Multi-Object Tracker}
Once we have learned a function to quantify the similarity between detections, we use these similarity values to merge redundant detections from the same objects.
We accomplish this with a modified version of the Hungarian algorithm~\cite{hungarian}.
The Hungarian algorithm provides a polynomial time solution to compute the minimum cost in a bipartite graph where each edge has a matching cost.
In each pair of consecutive frames, we construct a bipartite graph where each node represents a detection, and each edge connecting two nodes indicates the assignment cost for marking the two nodes as belonging to the same object.
By computing which assignments achieve the minimum sum of costs, similar detections as measured by the similarity network are most likely to be paired, and detections with greater pairing cost are less likely to be paired with one another.

One limitation of the Hungarian algorithm is that it always pairs as many nodes from the bipartite graph as possible.
For example, if one set in the graph has 5 nodes and the other set contains 4 nodes, the 4 pairings which minimize the sum of costs will be selected by the Hungarian algorithm.
This behavior is undesirable for our application, since it is possible for many signs to disappear from view while new signs appear, so pairing as many nodes as possible would result in nodes representing detections from different objects being paired.
We solve this problem with a simple modification to the algorithm.
If the distance computed between a pair of detections is greater than a cutoff threshold of 0.7, then they are forcibly split, meaning the detections are determined to correspond to separate objects.
The output of the tracker is a set of tracklets where each tracklet is a list of detections predicted to belong to the same sign.

\subsection{Geolocalized Sign Prediction}
The only remaining step in our pipeline is to condense the tracklets into sign predictions.
Since each detection in the tracklet is already predicted to belong to the same sign, we aggregate the list of GPS and class predictions contained within each tracklet into a singular sign prediction.
The simplest method is to predict a sign at the GPS coordinates and with the class from the last frame in the tracklet, which we refer to as the frame of interest (FOI) method.
A similarly simple approach is to take a weighted average of the GPS coordinates, and predict the class as being the mode of the detections in the tracklet.
Our third approach involves performing triangularization to condense the tracklets into sign predictions.
Finally, we can use the Markov Random Field model proposed in~\cite{MRF} to reduce the tracklets we have produced into sign predictions.

\section{Results}
\label{Results}

\begin{figure*}[!tp] 
    \centering
    \includegraphics[width=\linewidth,keepaspectratio]{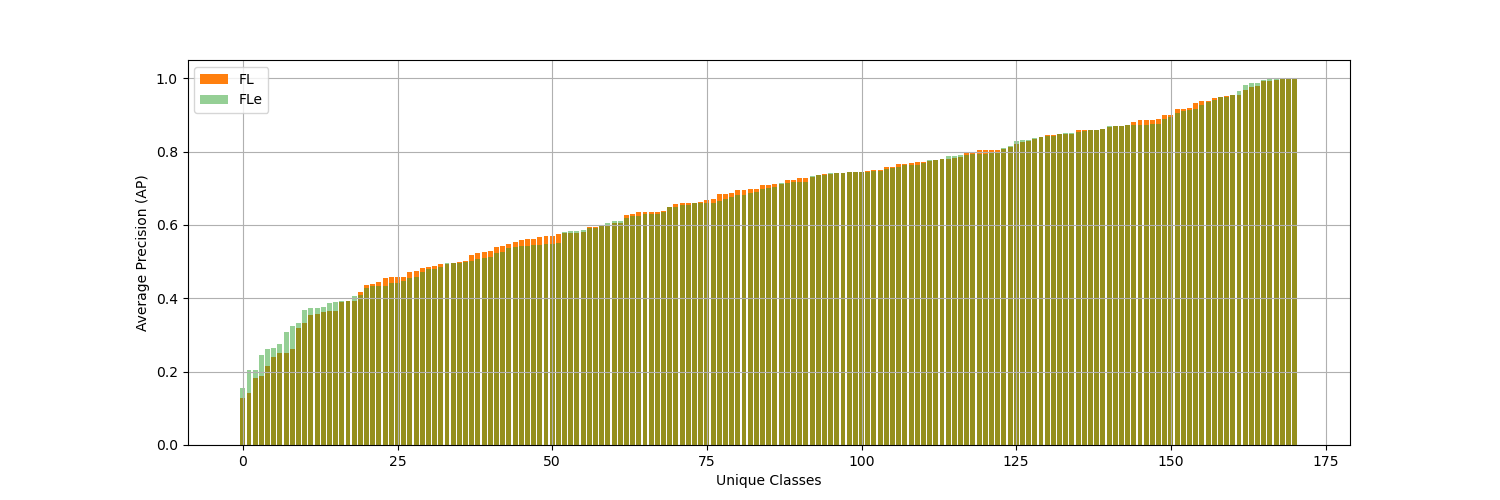}
    \caption{The underlying distribution of classes along with their \texttt{AP} scores for the best model found on the ARTS dataset using FL and FL$e$. 
    Each bar represents a unique sign class in the dataset and the height of the bar reflects the \texttt{AP} score of the given class. 
    Dark-green bars illustrate the proportion of classes where both FL and FL$e$ has approximately equivalent \texttt{AP} score. 
    Orange bars represents classes where FL scored better than FL$e$, while light-green bars refer to classes where FL$e$ scored a better \texttt{AP}. 
    FL$e$ demonstrates notably improved ability to increase the \texttt{AP} of hard classes (lower quartile proportion of classes), while preserving the performance on well-classified classes. }
    \label{fig:ap-dist}
\end{figure*}

\subsection{Object Detector Performance}
While the ultimate objective of our system is to perform object geolocalization, we first benchmark the performance of our object detection system to quantify its performance as an intermediary step when our system performs geolocalization.
We initialized our object detector with weights from a pre-trained model on the COCO dataset~\cite{lin2014coco}.
We kept the default optimization parameters provided by RetinaNet~\cite{lin2018retinanet} with the exception of increasing the initial learning rate to $1\mathrm{e}{-4}$.
We used smooth L1 loss on both the bounding box regression-subnet and the GPS-subnet.
The L1 loss for the GPS subnet is computed relative to the correct offset by transforming the annotated GPS coordinates to the local image coordinate system using the transformation outlined in Sec.~\ref{Methods}.
We use our custom focal loss function to train the classification subnet.
Our models were trained and tested on a workstation with a NVIDIA 2080ti GPU, as well as a computing cluster with NVIDIA Tesla V100 GPUs.
We report the mean average precision \texttt{mAP} evaluated at Intersection over Union \texttt{IoU=.5} on the Challenging version of the ARTS dataset.

\begin{table}[!ht] 
\centering
\begin{tabu} to \columnwidth {lC}
\toprule 
\textbf{Loss Function} & \textbf{ARTS v2 \texttt{mAP}$_{50}$} \\  \midrule \midrule
RetinaNet-50 (FL) & 69.9 \\ 
\textbf{RetinaNet-50 (FL$e$)} & \textbf{70.1} \\ \bottomrule 
\end{tabu} 
\caption{Traffic Sign Recognition Benchmark \texttt{mAP}$_{50}$ on the test subset of the ARTS v2~\cite{arts} dataset.} 
\label{tab:benchmark}  
\end{table}

Figure~\ref{fig:ap-dist} shows the distribution of classes and their \texttt{AP}'s.
Each unique sign is represented by a bar to indicate its \texttt{AP} score on the y-axis.
The orange color represents classes where FL scored better than FL$e$, while light-green refers to classes where FL$e$ scored better, and dark-green for approximately similar score.
Both loss functions are effective in increasing AP across class types, but FL$e$ demonstrates improved tail performance, with higher values for the $25^{th}$ and $75^{th}$ percentiles.
This result supports the effectiveness of FL$e$ in emphasizing low performing classes and ensuring that training gives more weight towards improving their \texttt{AP}.
Moreover, FL$e$ does not appear to have significantly decreased the \texttt{mAP} or the \texttt{AP} of classes that performed well with FL.
This suggests that FL$e$ is a sound compromise between promoting poorly performing classes and retaining the performance of easier classes.

\subsection{Object Detector GPS Prediction}
Each detection produced by the detector has a corresponding offset prediction which can be transformed to a GPS location using the previously established transformation. 
To quantify the performance of this component of our system, we compute the mean absolute error between the predicted distance produced by GPS-Retinanet and the ground-truth distance of the corresponding sign. 
Tho make construct an error metric easily interpretable by humans, we convert the absolute error between GPS locations to meters using the Haversine formula, which provides a accurate approximations at close distances. 
The \texttt{Haversine} formula is denoted as follows where $\delta$ is the relative distance, $\psi$ is latitude, $\lambda$ is longitude, and $R$ is the mean of earth's radius $6,371km$:

\begin{align*} 
    a & = \sin^2 \bigg( \dfrac{\Delta \psi}{2} \bigg) + \cos \psi_1 \cdot \cos \psi_2 \cdot \sin^2 \bigg( \dfrac{\Delta \lambda}{2} \bigg), \\
    \delta & = 2R \cdot atan2 \bigg( \sqrt{a}, \sqrt{1-a} \bigg). 
\end{align*}

\begin{figure}[!ht] 
    \centering 
    \includegraphics[width=\columnwidth,keepaspectratio]{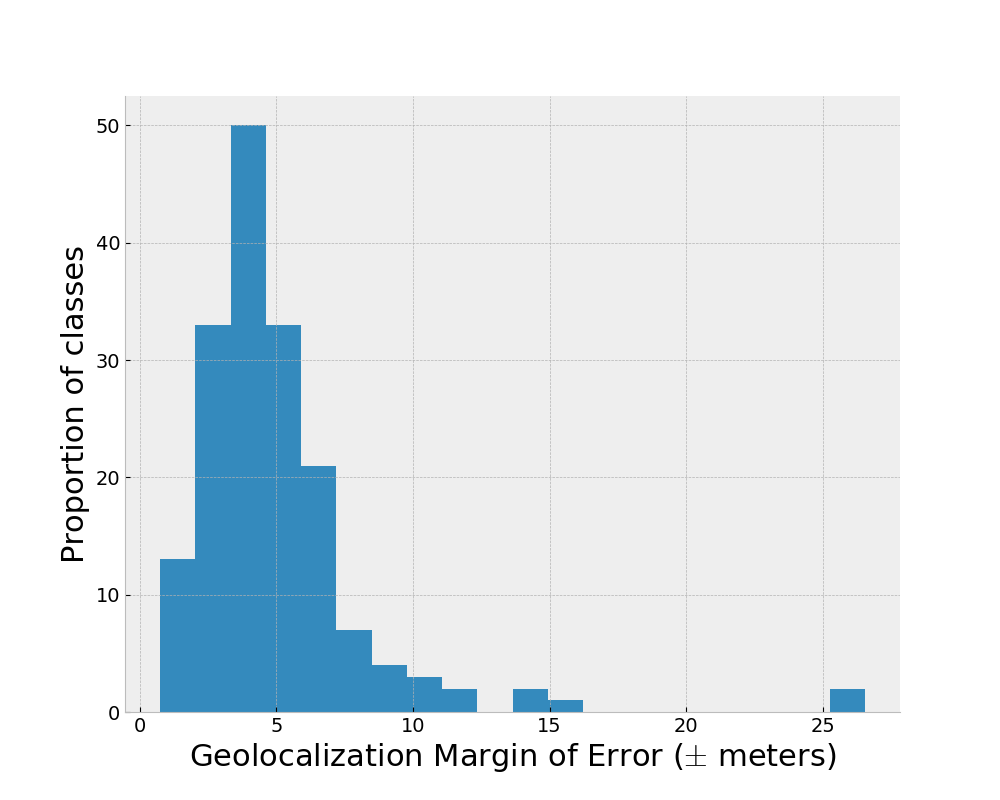}
    \caption{
        Average GPS testing error for each class.
        Our GPS-subnet scored a median MOE of ($\pm 5$) meters.
        We can see that the GPS-subnet can accurately estimate distance with a reasonably low margin of error.
    } 
    \label{fig:moe}  
\end{figure}

In Figure~\ref{fig:moe} we show the distribution of the mean prediction errors for each of the sign classes. 
We observe that most classes have mean predicted distances within 5 meters. There are a few outlier classes with much larger errors. 
This is largely a consequence of these signs appearing with low frequency in the dataset. 
Inspection of these difficult classes revealed they corresponded to signs which have a particularly broad distribution in their size, which is unsurprisingly challenging on a data set composed of images from a single camera.

\subsection{Similarity Network}
Next, we quantify the performance of the similarity network, which learns to predict value of 0 if the two input detections belong to the same physical sign and a value of 1 if the detections are from different signs.
Intuitively, the range of values from $0$ to $1$ can be interpreted as an abstract measure of ``distance'' between the two detections.
Since this network isn't performing classification, we can instead quantify its performance by measuring the error magnitude at different percentiles.
In Table~\ref{tab:similarity_heuristic}, each percentage indicates how often the network predicts a value at least as good as the listed error value.
We use $80\%$ of our detections for training the network, $10\%$ for validation, and the remaining $10\%$ is reserved for testing.

\begin{table}[!ht] 
\centering
\begin{tabu} to \columnwidth {Xc}
\toprule 
\textbf{Percentile} & \textbf{Error \%} \\  \midrule \midrule
50 & 0.0165 \\ 
75 & 0.1195 \\
90 & 0.3846 \\ 
95 & 0.6106 \\
97 & 0.7436 \\ 
98 & 0.8064 \\
99 & 0.8844 \\ \bottomrule 
\end{tabu} 
\caption{A table showing the distribution of prediction errors. Each percentile indicates the percent of errors that are at worst equal to the listed error value.} 
\label{tab:similarity_heuristic}  
\end{table}

\subsection{Tracker}
The objective of the tracker is to collapse down the detections produced by RetinaNet into geolocalized sign predictions.
Object geolocalization using deep learning is a new a growing field.
There are yet to be any universally accepted performance metrics, especially since performance in this domain is particularly sensitive to the difficulty of the dataset.
The goal of our performance evaluation is to quantify how well the physical sign predictions match up with the annotated physical signs distinguished in the ARTS dataset by their ID.
Specifically, we define a true positive as when the tracker predicts a sign that correctly matches to a real sign with 15 meters.
We define a false negative as a circumstance where there exists a real sign, but the tracker fails to generate a corresponding prediction.
Lastly, we define a false positive to be when the tracker predicts a sign, but no real counterpart exists.
An ideal tracker should achieve as many true positives as possible, while minimizing the count of false negatives and false positives.

\begin{table}[ht!]
\centering
\begin{tabu} to \columnwidth {lCCC}
\toprule 
\multicolumn{4}{c}{\textbf{End-to-End Performance}} \\ \midrule
\textbf{Year} & \textbf{2012} & \textbf{2013} & \textbf{2014} \\ 
\midrule \midrule
True Positives & 264 & 3170 & 3179  \\ 
False Negatives & 176 & 604 & 842  \\ 
False Positives & 67 & 826 & 1581  \\ \bottomrule
\end{tabu}
\caption{End-to-end system performance benchmark.} 
\label{tab:end_to_end}
\end{table}

\begin{figure}[!ht] 
    \centering 
    \includegraphics[width=\columnwidth,keepaspectratio]{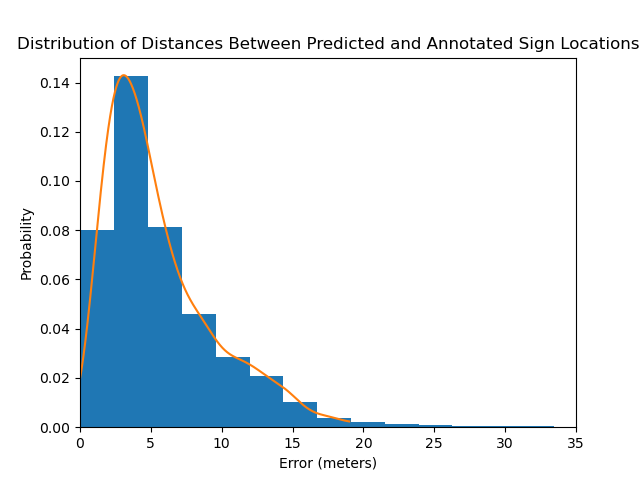}
    \caption{
        A probability distribution of GPS errors between the predicted geolocalized sign coordinates and the actual coordinates from the annotations.
    } 
    \label{fig:geolocalization_error}  
\end{figure}

In Table~\ref{tab:end_to_end}, we show the number of true positives, false negatives, and false positives during different years containing different road segments.
The data for geolocalization is divided into years in which it is gathered, and each year contains road segments that the tracker steps through to perform geolocalization.
The 2012 and 2018 data contain mostly driving through towns, and are hence the most challenging due to all the signs and extra objects present.
2014 contains many rural segments, and is thus not quite as challenging.
The 2013 data is composed mostly of highways, and is therefore the least challenging group of road segments.
The table also provides the distribution of GPS errors between each geolocalized sign marked as a true positive and its corresponding annotated sign.

\subsection{Comparison to other Methods}
Comparison to existing geolocalization techniques is challenging due to the lack of standardized evaluation metrics and varying structure to datasets.
We believe our dataset is the most representative of data encountered by geolocalization systems in the real world, however this also limits the comparisons we are capable of performing.
For example, it is impossible for us to compare our results to \cite{uber}, since their method uses 5D pose data which is unavailable in our dataset.
Many other tracking methods do not transfer well to our problem either due to not being designed to deal with the very low frame rate or the broad and sparse class distribution contained by the dataset.
We therefore perform two sets of comparisons.

We run some classic tracking algorithms on the detections produced by RetinaNet, to evaluate weather traditional tracking methods are capable of performing in a low frame rate environment.
In Table~\ref{tab:tracker_comparisons}, we run multiple trackers and find that they are essentially completely unable to track on this dataset.
The reason is they are unable to locate the large distances the objects jump between frames, and hence are unable to merge redundant detections corresponding to the same sign.
This is justifies why we need to train the distance network to compute a learned heuristic that represents the similarity of two signs, since it makes to possible to detect detections from the same sign.

\begin{table*}[ht!]
\centering
\begin{tabu} to \textwidth {lCCCCC}
\toprule 
\multicolumn{6}{c}{\textbf{Geolocalization Performance Comparisons}} \\ \midrule
\textbf{Tracking Method} & \textbf{True Positives} & \textbf{False Negatives} & \textbf{False Positives} & \textbf{Mean GPS Error} & \textbf{STD GPS Error}\\ 
\midrule \midrule
Frame of Interest & 6677 & 2759 & 1558 & 5.85 & 4.40 \\ 
Weighted Average & 6670 & 2751 & 1565 & 5.81 & 4.38 \\ 
Triangularization & 6079 & 3000 & 1918 & 6.67 & 4.33 \\ 
MRF & 6677 & 4379 & 2156 & 6.57 & 4.98 \\ 
\midrule
\bottomrule
\end{tabu}
\caption{Performance comparison using different methods to geolocalize signs from detections.} 
\label{tab:geolocalization_comparison}
\end{table*}

While there are not directly analogous state-of-the art approaches to compare to, we can compare the geolocalization performance of difference techniques on our tracklets.
Each algorithm receives as input each sequence of detections formed by using the similarity scores between detections computed by our tracker network as input to the hungarian algorithm.
Each sequence, or ``tracklet'' contains a list of detections that are computed as similar to one another, meaning they likely belong to the same sign.
We compare the ability of three algorithms to geolocalize these tracklets.
The simplest method is to predict a sign at the GPS coordinates and with the class from the last frame in the tracklet, which we refer to as the frame of interest method.
A similarly simple approach is to take a weighted average of the GPS coordinates, and predict the class as being the mode of the detections in the tracklet.
Our third approach involves performing triangularization to condense the tracklets into sign predictions.
Finally, we use the Markov Random Field model propose in~\cite{MRF} to reduce the tracklets we have produced into sign predictions.
In Table~\ref{tab:geolocalization_comparison}, we see taking a simple weighted average is the most effective method of converting the tracklets into sign predictions.
It achieves the lowest GPS error, low standard deviation, and good scores for true positives, false negatives, and false positives.

\section{Discussion and Future Work}
\label{Discussion}

In this paper we present the ARTS v2 dataset, which serves as a traffic sign geolocalization benchmark.
Each sign annotation in ARTS v2 consists of a sign class, side of road indicator, and sign assembly indicator, and a unique sign identifier.

Using ARTS v2 we developed a two-stage pipeline that detects, tracks, and geolocalizes traffic signs.
Traffic sign detection is handled by GPS-RetinaNet, which predicts bounding box coordinates, sign class, and GPS offsets for each detected sign.
GPS-RetinaNet used a variant of the Focal loss during training to effectively handle the class imbalance present in ARTS.
Traffic sign tracking and geolocalization were handled using a learned metric network and a variant of the Hungarian algorithm.
This system can automatically detect, classify, and geolocalize traffic signs using only low frame rate video and camera pose information.

Since we focus on low frame rate video, our approach may not be suited for real-time tracking.
Future research should explore optimizations and tuning to facilitate this use case.
Geolocalization accuracy may be improved by the use of alternative loss functions, since the smooth L1 loss penalizes the network harshly for classes it is struggling with.
The roads from which our dataset is gathered are surveyed during multiple years, so it may be possible to aggregate data accross different years to enhance results.
Improved hardware such as LIDAR could certainly enhance performance, but with the downside of increased cost.
Proper evaluation metrics for geolocalization is also a topic deserving of future research.
The noise introduced to GPS coordinates due to both equipment error and annotation inconsistencies limits the capability of GPS to serve as a ground truth.
Future work could also experiment with how to better distinguish between signs with similar visual features and locations during tracking, as these objects have the fewest distinguishing features.
Finally, it may be fruitful to develop an end-to-end system, rather than our two-stage approach, to simplify the training and tuning process.

\section{Acknowledgments}
\label{Acknowledgements}
Computations were performed using the Vermont Advanced Computing Core supported in part by NSF award No. OAC-1827314.
We thank Josh Minot and Fayha Almutairy for their contributions, discussions and feedback on this project.
This work would not have been possible without great collaboration between the Vermont Artificial Intelligence Lab and VTrans.
We would like to thank Rick Scott and Ken Valentine for championing this project.
We also wish to thank the reviewers for their insightful comments.

\section{Funding}
\label{Funding}
This research was funded by the Vermont Agency of Transportation.


\appendix

\section{Appendix}
\label{appendix}
The Focal Loss was first introduced in~\cite{lin2018retinanet} to address the challenge of overwhelming the loss value of rare classes with many easy classes during training for datasets with unequally distributed samples.
One of the most crucial properties of the FL is the basic idea of down-weighting the loss of easy (well-classified) classes in favor of focusing the training on the hard classes in the dataset.
The Focal Loss is defined as:
\begin{align}
    \textnormal{FL}(p_t) & = -(1 - p_t)^{\gamma} \log(p_t).
\end{align} 
The focusing parameter $\gamma$ acts as a modulator to fine-tune the effect of down-weighting the loss of easy classes.
\cite{lin2018retinanet} noted that $\gamma = 2$ works well, since it does not penalize maintains acceptable performance on easy classes while noticeably improving performance on hard classes.
In our experiments, however, we found that fixing the focusing parameter value for all classes results in an unintended effect in which the loss value of a wide range of classes starts to get down-weighted prematurely not allowing them to achieve better average precision in a reasonable amount of time.
In other words, FL increasingly down-weights the loss value of all classes once their probability $p_t$ surpasses $.3$, which one can argue that it is too low to consider as a threshold for `well-classified' classes. 

We propose a modification to the Focal Loss that replaces $\gamma$ in the original definition by an adaptive modulator. 
We define the new focusing parameter as: 
\begin{align}
    \Gamma & = e^{(1 - p_t)}, \\
    \textnormal{FL}e(p_t) & = -(1 - p_t)^{\Gamma} \log(p_t).
\end{align} 

\begin{figure*}[!h] 
    \centering 
    \includegraphics[width=.95\textwidth]{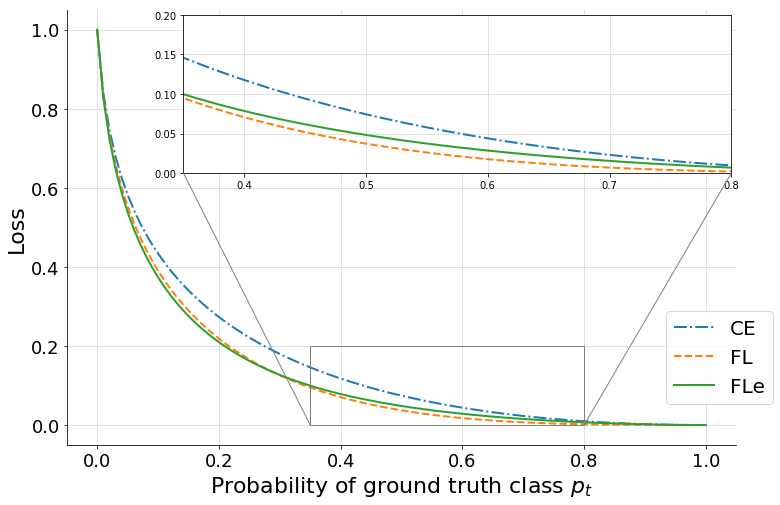}
    \caption{
        Our modified Focal Loss function (FL$e$) compared with FL ($\gamma$ = 2), and cross entropy (CE).
        FL$e$ introduces an adaptive exponent to the original FL~\cite{lin2018retinanet}.
        This effectively changes the underlying distribution of classes in regards to their \texttt{AP}s and promotes some of the poorly classified classes to a better score while preserving the performance of well-classified classes.
    } 
    \label{fig:loss}  
\end{figure*}  

For convenience, we will refer to our new definition of Focal Loss as (FL$e$).
FL$e$ introduces two new properties to the original definition.
It dynamically fine-tunes the exponent based on the given class performance to reduce the relative loss for well-classified classes maintaining the primary benefit of the original FL.
Figure~\ref{fig:loss} directly compares FL with FL$e$, highlighting that FL$e$ (shown in green) crosses over $FL_{\gamma=2}$ (shown in orange) around ($p_t = .3$).
As $p_t$ goes up from $.3 \to 1$, FL$e$ starts to shift up slowly ranging in between FL and Cross Entropy CE (shown in blue). 

In practice, this allows us to ultimately define `well-classified' classes as ($p_t > .7$) instead of ($p_t > .3$) in the original definition.
In other words, FL$e$ reduces the loss down-weighting effect on classes when their $p_t$ values are in the range ($.3 \geq p_t \geq .7$) while still focusing on hard classes.
This results in slightly improved performance that manifests at the beginning of the training and continues throughout the process till both FL and FL$e$ converges at a similar \texttt{mAP}.
However, FL$e$ will have a slightly lower standard deviation as more classes will cluster around \texttt{mAP} whereas FL will have a greater spread of \texttt{AP}'s per class.

\begin{table*}[ht!]
\centering
\begin{tabu} to \textwidth {lCCC}
\toprule 
\multicolumn{4}{c}{\textbf{Tracker Performance Comparisons}} \\ \midrule
\textbf{Tracker} & \textbf{True Positives} & \textbf{False Negatives} & \textbf{False Positives} \\ 
\midrule \midrule
Boosting & 8062 & 173 & 24425  \\ 
MIL & 8068 & 167 & 24130  \\ 
KCF & 8061 & 175 & 25812  \\ 
TLD & 8055 & 180 & 21903  \\ 
MedianFlow & 8054 & 181 & 20834  \\ 
GoTurn & 8049 & 186 & 22203  \\ 
MOSSE & 8042 & 193 & 21422  \\ 
CSRT & 8061 & 174 & 23052 \\ 
\midrule
\textbf{ARTS} & \textbf{6677} &\textbf{2759} & \textbf{1558} \\ 
\bottomrule
\end{tabu}
\caption{Performance using different trackers to geolocalized detections from GPS-Retinanet.} 
\label{tab:tracker_comparisons}
\end{table*}

 \bibliographystyle{elsarticle-num} 





\end{document}